\documentclass[10pt]{article} 
\pdfoutput=1 
\usepackage[preprint]{tmlr}

\usepackage{amsmath,amsfonts,bm}

\def\eqref#1{equation~\ref{#1}}

\def\1{\bm{1}}

\DeclareMathAlphabet{\mathsfit}{\encodingdefault}{\sfdefault}{m}{sl}
\SetMathAlphabet{\mathsfit}{bold}{\encodingdefault}{\sfdefault}{bx}{n}

\newcommand{\rhoFull}{0.9339}
\newcommand{\rhoAbstract}{0.934}
\newcommand{\rhoIntro}{0.93}
\newcommand{\rhoCIpctLo}{0.8940}
\newcommand{\rhoCIpctHi}{0.9673}
\newcommand{\rhoCIbcaLo}{0.8970}
\newcommand{\rhoCIbcaHi}{0.9698}
\newcommand{\rhoHone}{0.5935}
\newcommand{\rhoHtwo}{0.5313}
\newcommand{\rhoHoneShort}{0.59}
\newcommand{\rhoHtwoShort}{0.53}
\newcommand{\nCkpts}{114}
\newcommand{\nRuns}{20}
\newcommand{\nBlocks}{10}
\newcommand{\sigmaCkpt}{0.1334}
\newcommand{\sigmaCont}{0.00945}

\newcommand{\rTwoA}{0.92}
\newcommand{\rTwoR}{0.89}
\newcommand{\rTwoG}{0.91}
\newcommand{\rTwoF}{0.94}
\newcommand{\effRank}{4.48}
\newcommand{\effRankCubic}{4.45}
\newcommand{\resCosine}{0.52}
\newcommand{\resLogitGap}{0.61}
\newcommand{\resAdamW}{0.80}

\newcommand{\oosCorrLo}{0.33}
\newcommand{\oosCorrHi}{0.46}
\newcommand{\oosRsqLo}{0.095}
\newcommand{\oosRsqHi}{0.175}
\newcommand{\oosPLo}{0.0005}
\newcommand{\oosPHi}{0.005}

\newcommand{\Greal}{0.1606}
\newcommand{\GrealPct}{16.1\%}
\newcommand{\GrealShort}{0.161}
\newcommand{\shamMedian}{0.1294}
\newcommand{\shamQfive}{0.2191}
\newcommand{\shamFrac}{30\%}
\newcommand{\Gexcess}{0.031}
\newcommand{\spearA}{0.67}
\newcommand{\spearR}{0.52}
\newcommand{\spearG}{0.52}

\newcommand{\alphaCross}{1}
\newcommand{\alphaEffMed}{0.092}
\newcommand{\alphaThresh}{0.710}
\newcommand{\ratioFold}{7.7}
\newcommand{\GcurveZero}{0.1606}
\newcommand{\GcurveQuarter}{0.1424}
\newcommand{\GcurveHalf}{0.1763}
\newcommand{\GcurveOne}{0.2721}
\newcommand{\GcurveTwo}{0.3458}
\newcommand{\GcurveFour}{0.3253}
\newcommand{\excHalf}{0.197}
\newcommand{\excOne}{0.016}

\newcommand{\panelMedARG}{0.1101}
\newcommand{\panelMedARGShort}{0.11}

\newcommand{\panelMedCorr}{0.1139}
\newcommand{\panelMedCorrShort}{0.114}
\newcommand{\panelRealCorr}{0.0505}
\newcommand{\panelRealCorrShort}{0.051}
\newcommand{\panelMedNoise}{0.0306}
\newcommand{\panelMedNoiseShort}{0.03}

\newcommand{\manifestHash}{8623522baf8c71f94d4b58bec92046d4e3ccd6cba176f7f16420d42f1e9bb6cc}

\usepackage{hyperref}
\usepackage{url}
\usepackage{graphicx}
\usepackage{booktabs}
\usepackage{amsmath}

\title{Response-Relevant Information Is Not Sufficient for\\Neural Training-State Equivalence}

\providecommand{\repourl}{\url{https://github.com/ClevixLab/training-state-equivalence}}
\def\ackstext{}
\IfFileExists{authors_private.tex}{
\author{\name Truong Xuan Khanh \email khanh@clevix.vn \\
      \addr H\&K Research Studio, Clevix LLC, Hanoi, Vietnam
}
\renewcommand{\repourl}{\url{https://github.com/ClevixLab/training-state-equivalence}}
\def\ackstext{}
}{%
\author{\name Anonymous authors \email \\
      \addr Paper under double-blind review}}

\def\month{MM}
\def\year{YYYY}
\def\openreview{\url{https://openreview.net/forum?id=XXXX}}

\begin{document}

\maketitle

\begin{abstract}
Response-relevant information is not sufficient evidence that two neural
networks are in the same training state.
We define training-state equivalence operationally: two checkpoints are
equivalent, with respect to an intervention family and a horizon, if they
respond alike when training is perturbed and continues. On fresh runs of a
one-layer transformer learning modular addition, response to a weight-decay
pulse is highly reproducible relative to the differences between checkpoints
($\rho_{\mathrm{state}} = \rhoAbstract$ at 500 steps), establishing the
target as measurable.
Fifteen descriptors of weight spectra, representation geometry and optimiser
geometry then carry genuine within-run response-relevant information beyond
weight norm and loss on a development dataset, predicting held-out response
deviation across seed blocks. But their apparent \GrealPct{} advantage at
matching checkpoints by future response does not separate from a conditional
placebo preserving each descriptor's fitted dependence on the conventional
observables, and a calibrated positive control makes simple insensitivity an
unlikely explanation.
Information a decoder can extract need not be organised in the pairwise
geometry that similarity judgments require.

\end{abstract}

\section{Introduction}\label{sec:intro}
Similarity between neural networks asks whether two networks are alike now.
A different question is whether they are alike in what they will do next.
Two checkpoints on two training runs can agree on loss, on weight norm and
on the geometry of their representations, and still be at different points
in whatever process determines how training proceeds from there.

That distinction turns on a requirement that is easy to state and easy to
conflate with a weaker one. Predicting an attribute from a representation
needs a decoder: some function of the representation must recover the
attribute. Deciding that two examples are equivalent with respect to that
attribute needs something else --- the pairwise geometry of the
representation must order examples by it. A representation can satisfy the
first requirement and fail the second, because the distances that decide
which examples are near each other may be dominated by directions carrying
nothing about the attribute. This paper is a concrete instance of that gap,
in a setting where the attribute is a network's future behaviour.

Existing answers characterise the present object. Weight-space measures ask
whether two parameter vectors are equivalent under the symmetries of the
architecture; function- and representation-space measures ask whether two
networks compute or encode the same thing. None is a claim about future
behaviour, and one line of work argues explicitly that behavioural
comparisons cannot substitute for them, being evaluated on whatever slice of
input space is at hand \citep{nissengonzalez2026tensor}. A separate
literature perturbs networks and reads structure from the response, but uses
response to characterise a single network or training trajectory rather than
to test whether two independently trained networks are interchangeable.

We take the second question directly, and treat similarity of future
intervention response as an operational criterion for training-state
equivalence: relative to a class of interventions and a horizon, two
checkpoints provide stronger evidence of sharing a state when their
reproducible responses are more similar. This is a criterion for evidence,
not a formal equivalence relation, and it inherits an operational scope ---
it says nothing about interventions or horizons at which it was not
measured. Training state is not defined here as intervention response
itself; reproducible response is used as one operational observable of
state. If two checkpoints are claimed to occupy the same state relative to
a specified intervention family and horizon, materially different future
responses are evidence against that claim, and agreement under one
intervention is local evidence, not a universal equivalence relation. Its
first requirement is empirical. If the same checkpoint,
perturbed identically twice, responds differently by as much as two
different checkpoints do, then there is nothing here to identify.

It is not empty. On fresh runs, with control and intervention branches
sharing random numbers so that they differ only in the perturbation, the
response to a weight-decay pulse is reproducible across independent
continuations to a degree that leaves differences between checkpoints
clearly resolvable: $\rho_{\mathrm{state}} = \rhoIntro$ at a short horizon,
falling as continuation noise accumulates over longer ones. Future response
is something a checkpoint has.

The natural next step is to look for it in the network. Fifteen descriptors
of each checkpoint --- weight spectra, representation geometry, optimiser
geometry --- carry structured residual variation beyond weight norm and
loss. That residual is not merely leftover variance. After removing each
run's mean from both the descriptors and the response, deviations in one
predict deviations in the other, out of block, across all five reported
ridge penalties. By conventional evidence, the identification problem would
appear solved.

It is not, and the reason is the distinction we began with. The result above
is \emph{within} a run: it says the descriptors know where a checkpoint sits
relative to its own trajectory. Equivalence is only useful if it compares
checkpoints whose trajectories differ. When the same descriptors are asked
to pick, for each checkpoint, a partner from an independently trained run
that will respond as it does, their apparent advantage over the conventional
observables does not survive a control. Permuting the descriptors' residual
between runs while keeping their fitted dependence on the conventional
observables intact leaves a placebo that the observed \GrealPct{}
improvement sits inside: \shamFrac{} of placebo runs match or exceed it. A
calibrated positive control detects linearly aligned signal only at a scale
substantially above the descriptors' measured aligned component, making
simple insensitivity to that signal class an unlikely explanation. A
synthetic arm locates most of the placebo's own level in the descriptors'
correlation with the conventional observables.

\textbf{Response-relevant variation within training runs is not sufficient
to establish training-state equivalence across runs.} Each trajectory may
admit a serviceable local coordinate; that does not make the coordinates of
different trajectories commensurable.

We make three contributions, at two levels of evidence.

\textbf{First}, on fresh confirmatory runs, short-horizon intervention
response is a highly reproducible checkpoint-level property relative to
independent continuation variability, making an intervention-based notion of
training state measurable rather than merely definable.

\textbf{Second}, on a pooled development dataset, internal descriptors carry
within-run response-relevant information beyond the conventional
observables --- surviving seed-block-held-out prediction against a
structured permutation null --- yet that information does not deliver a
cross-run response-equivalence geometry under the transport criterion we fix
in advance.

\textbf{Third}, a structured conditional placebo isolates a substantial
apparent advantage attributable to a representation's fitted dependence on
conventional observables, and a calibrated positive control shows that the
resulting negative conclusion is not readily attributed to insensitivity at
the tested signal scale.

We develop the analysis on a pilot cohort, freeze it, and evaluate the
reproducibility claim on fresh runs; the descriptor analyses are
development-dataset results and are labelled as such. All descriptor
analyses use the prespecified $h=500$ response, where the confirmatory
experiment established the highest reliability, and conclusions about
descriptor relevance and transport apply at that horizon. Everything here
concerns one task, one architecture, one intervention family, one descriptor
set and one transport criterion.

\section{Related work}\label{sec:related}

\subsection{What existing similarity measures establish, and what they do not}

Two networks can be called alike because their weights are alike, because
their functions are alike, or because they will behave alike. Much of the
neural-network similarity literature has concentrated on the first two.

\citet{nissengonzalez2026tensor} set out three criteria a similarity measure
should meet --- independence from a particular dataset, invariance to the
symmetries of the parameterisation, and tractability at scale --- and argue
that behavioural similarity fails the first, because it compares
input--output pairs on whatever slice of input space it is evaluated on and
is not guaranteed to generalise outside it. Their measure is computed from
weights alone and meets all three, at the cost of requiring tensor-based
architectures; they state plainly that contemporary networks do not appear
to admit a measure that tractably satisfies all three. Their setting
overlaps ours closely --- modular addition, AdamW with weight decay,
checkpoints through training, Fourier structure in the embedding --- and
their pairwise similarity matrices compare checkpoints along a single run.

Our object differs in three respects. We compare networks trained
independently, from different seeds and different hyperparameters, rather
than points on one trajectory. We define likeness by future behaviour under
a training intervention rather than by present weights. And we work with a
standard transformer, where their measure does not apply.

Their criticism applies to us and we accept it rather than answer it: a
criterion built on measured response is behavioural, and it speaks only
about the interventions and horizons at which it was measured. The two
notions answer different questions; neither supersedes the other.

\citet{kwok2025butterfly} perturb weights --- Gaussian noise, a single
weight, or a sampled batch-gradient direction --- at a chosen time and
continue training deterministically, measuring divergence by parameter
distance, loss barrier, permutation-aligned distance and representational
similarity. Sensitivity is high early in training and falls sharply, which
is already a statement that response depends on training state. Their design
removes all training noise in order to isolate the perturbation, so
continuation-to-continuation variability is not a quantity their protocol
produces; the reproducibility question we ask requires that variability to
exist and be measured against.

Mode-connectivity work asks whether independently trained solutions can be
joined by low-loss paths in parameter space
\citep{garipov2018loss, draxler2018essentially}. Our criterion is
orthogonal: two checkpoints may lie on a connected low-loss set yet differ
in their response to subsequent optimisation perturbations, and conversely
response similarity does not imply a low-loss connecting path.

\subsection{What is already known about reading structure from response}

Perturbing a system and reading its structure from the response is an active
programme in interpretability, and we do not claim the framing.

\citet{baker2026structural} define the susceptibility of an observable to a
perturbation of the data distribution as the derivative of its posterior
expectation, show by a fluctuation--dissipation argument that it equals a
posterior covariance, and estimate it from the unperturbed posterior by
local SGLD sampling. Assembled into a response matrix, these separate
functional modules in a small transformer. The programme continues by
clustering per-token susceptibility vectors into interpretable groups
\citep{gordon2026spectroscopy}, by visualising how those vectors organise
over training \citep{wang2025embryology}, and by inverting the map to find
the data perturbation that produces a desired structural change
\citep{elliott2026susceptibilities}.

The mechanism differs from ours in three ways that matter. Their
perturbation is of the data distribution; ours is of the optimiser. Their
response is an expectation under a localised posterior around the current
parameters; ours is what the network does when training actually continues.
Their observables are localised on components; ours is global generalisation
performance at a horizon.

One question their framework does not pose is whether the response is
reproducible. Their susceptibilities are posterior expectations estimated by
sampling, so there is no independent continuation whose variability could be
compared against the spread between checkpoints. That comparison is the
first thing we measure.

\subsection{Why encoded information is not evidence of use}

The distinction our fourth result depends on has been drawn clearly before.

\citet{elazar2021amnesic} argue that probing accuracy shows a property can
be extracted from a representation but gives no evidence for or against the
model using it, and instead remove the property by iterative nullspace
projection and measure the behavioural consequence. They conclude that
conventional probing performance is not correlated with a property's
importance to the task. \citet{ravfogel2021counterfactual} add the
complementary methodological requirement: a good intervention alters one
intended factor, leaves the rest as intact as it can, and evaluates the
behavioural change against a directional prediction derived from linguistic
theory. \citet{belinkov2022probing} organises the recurring criticisms ---
comparative baselines, probe complexity, task and dataset choice, and the
gap between correlation and causation --- and points to intervention-based
designs wherever the claim concerns use rather than extractability.

We are inside this line rather than opening it. What differs is the target.
In that literature the behaviour of interest is the model's current output;
here it is what the network does next when training is perturbed. Turning
``present but not used'' into a question about training-state equivalence,
where the downstream use is matching two checkpoints for future response, is
the move we make.

The field has produced several distinct answers to the confound that a probe
may succeed on its own capacity rather than on the representation.
\citet{hewitt2019designing} construct control tasks by assigning each word
type an independently sampled random behaviour --- random across types,
deterministic for every token of a type --- and define selectivity as
linguistic-task accuracy minus control-task accuracy; they use selectivity
to guide probe design and interpretation, including the choice of probe
complexity and regularisation. \citet{voita2020information} show that such
probe-design constraints are avoidable: with minimum description length,
accuracy on the real and control tasks can be close --- 97.5 against 91.9 at
one layer --- while codelength separates them by almost six-fold, 85 against
470 kbits. The ``Rand'' control of amnesic probing \citep{elazar2021amnesic}
takes a third route, removing the same number of directions as the real
intervention but choosing them at random.

Each addresses a different confound: probe capacity and memorisation; the
effort a probe needs, whether measured as model cost or as sample
efficiency; and the dimensionality of what was removed. These target
different confounds from the one that arises here --- the apparent
downstream advantage associated with a representation's fitted relationship
to conventional covariates, which appears when a rich representation is
compared against a conventional baseline on a matching task.

\subsection{What our control does, and what it borrows}

The control we use follows an established principle. \citet{candes2018panning}
introduced the conditional randomisation test; \citet{berrett2020conditional}
developed the conditional permutation test, in which the values of $X$ are
permuted non-uniformly so as to respect the dependence between $X$ and $Z$,
and proved that type-I inflation is bounded by the error in approximating
the conditional distribution of $X$ given $Z$. They note that several
earlier proposals amount to binning $Z$ and permuting within bins, which is
close to the cell structure we work in. \citet{kim2022local} analyse
within-bin permutation procedures and show that their validity is
approximate, depending on the smoothness of the conditional structure and on
bin construction. This provides an additional reason to calibrate our
structured placebo empirically rather than invoke exact
conditional-randomisation guarantees.

Our placebo borrows this principle but is not an implementation of the exact
procedures those guarantees cover: the conditional distribution is
approximated by a fitted low-order regression, the permutation is further
constrained by seed-bundle and availability structure, and the statistic is
the outcome of a downstream fitting-and-selection pipeline. We therefore
calibrate its behaviour empirically --- including how much injected signal
it can detect --- rather than invoking those guarantees.

\section{Setting and methods}\label{sec:methods}

\subsection{Task, architecture and training}

Modular addition, $(a+b) \bmod p$ with $p=97$, as a classification problem
over the 97 residues. Inputs are the two operands as tokens; the target is
the sum. The 9{,}409 pairs are split 50/50 into training and test sets by a
fixed permutation shared across every run, so that ``test loss'' means the
same held-out set everywhere.

The model is a one-layer transformer encoder: token embedding of width 128,
a learned positional parameter for each of the two positions, four attention
heads, a feed-forward width of 512, pre-normalisation, no dropout, and a
linear readout from the final position to 97 logits.

Training uses AdamW with learning rate $10^{-3}$, $\beta=(0.9, 0.98)$, batch
size 512 drawn with replacement, for 30{,}000 steps. Two hyperparameter
conditions differ only in weight decay: $\lambda=0.5$ and $\lambda=1.0$.
Both reach test accuracy 1.0 within the budget in every run.

\subsection{Cohorts and evidence regimes}

Five paired seed blocks were trained first and used to develop the design:
to fix the checkpoint-selection rule, to inspect the geometry of the
trajectories, and to choose the conditioning set. Ten fresh paired seed
blocks were then trained for the confirmatory response-reproducibility
experiment. The two regimes are kept distinct and are labelled wherever a
result is reported.

\textbf{R1 uses only the ten fresh blocks.} The selection rule, the
endpoint, the horizon and the estimand were all fixed before those runs
existed, so this is confirmatory evidence on a held-out cohort.

\textbf{R2 to R5 use a pooled fifteen-block dataset} comprising the original
five development blocks and the ten used in R1. Their
leave-one-seed-block-out procedures give out-of-block prediction within that
pooled dataset; they do not constitute independent replication on a
study-level holdout. This matters asymmetrically: R4 and R5 are
falsification and calibration analyses, which do not carry the optimism risk
that a positive finding would, whereas R3 is a positive finding computed on
data that shaped the choice of descriptors, conditioning variables and state
space.

Seeds are shared across conditions: run $(s, \lambda=0.5)$ and run
$(s, \lambda=1.0)$ start from the same initialisation and consume the same
batch sequence, and diverge only through weight decay. The pair is therefore
treated as one \textbf{seed block} throughout, and the seed block, not the
run, is the independent unit for every interval and every permutation.

We use ``preregistered'' only for choices fixed before any data were seen.
Choices fixed on the pilot cohort and then held unchanged are described as
frozen after development and before confirmatory evaluation.

Checkpoints are chosen by a mechanical rule on a $3\times 3$ grid over
(weight norm, $\log_{10}$ training loss), fixed before the confirmatory
cohort was trained, so that no checkpoint enters the analysis because of how
it responded; the full rule, including the first-passage restriction and
support conditions, is given in Appendix~\ref{app:selection}. On the
confirmatory cohort six of nine cells qualify, yielding \nCkpts{}
checkpoints in the qualifying cells.

\subsection{Intervention and continuation}

The intervention is a weight-decay pulse: for 200 steps from the checkpoint,
$\lambda$ is multiplied by three, then restored. Everything else is
unchanged.

Each checkpoint is continued three times. Within a replicate, a control
branch and an intervention branch start from the same checkpoint, restore
the same optimiser state, are seeded identically, and consume an identical
precomputed batch sequence; they differ only in the pulse. The continuation
seed is a hash of (checkpoint identity, replicate), so replicates are paired
within a checkpoint but independent across checkpoints --- a single global
seed schedule would give every checkpoint the same batch sequence for a
given replicate, and the between-checkpoint variance estimator assumes
independent groups. Baseline hyperparameters are read from the restored
optimiser rather than from the configuration file, since the two conditions
differ in $\lambda$.

Two properties are enforced rather than assumed. \textbf{Bitwise
determinism:} deterministic kernels are required, and before measurement a
control branch is run twice from the same seed on three checkpoints spanning
early, middle and late training; the two runs must agree bitwise in model
tensors, optimiser state and raw logits. Accuracy agreement is not
sufficient: two diverged trajectories can round to the same accuracy.
\textbf{Storage disjointness:} every branch asserts that no tensor in its
model or optimiser state shares storage with the checkpoint it was restored
from. Implementation details are given in Appendix~\ref{app:determinism}.

\subsection{Endpoint}

The primary endpoint is the paired difference in $\log_{10}$ test loss at
horizon $h=500$ steps:
\begin{equation*}
D \;=\; \log_{10}\max(L^{I},\varepsilon) \;-\; \log_{10}\max(L^{0},\varepsilon),
\qquad \varepsilon = 10^{-8}.
\end{equation*}
Test loss rather than accuracy: at these horizons accuracy is bimodal, with
61\% of checkpoints below 0.01 and 14\% above 0.99 at $h=500$, so a paired
difference in accuracy would be zero for most pairs --- not because the
intervention had no effect but because the measure cannot resolve it. The
floor $\varepsilon$ never binds; the smallest observed test loss is
$3.8\times 10^{-7}$. Horizons 1000 and 2000 are recorded and reported as
secondary. All descriptor analyses use the prespecified $h=500$ response,
where the confirmatory experiment established the highest response
reliability; conclusions about descriptor relevance and transport therefore
apply to this horizon.

\subsection{Descriptors and conventional observables}

Twenty scalars are extracted per checkpoint, in four families of five.

\textbf{A, weight spectra:} singular-value statistics --- spectral entropy,
stable rank, top-1 and top-5 energy, and $\sigma_1$ divided by the trace ---
averaged with equal weight per matrix over the token embedding, the
separated Q, K, V and output projections, both feed-forward matrices, and
the unembedding. Parameter-count weighting would make this family
essentially the feed-forward spectrum, since the two feed-forward matrices
hold about 70\% of the eligible parameters while the embedding holds about
7\%.

\textbf{R, representation and function,} on a fixed evaluation probe of 512
held-out pairs: effective rank of the final-layer activation covariance,
top-5 principal-component share, representational anisotropy, predictive
entropy, and the dispersion of the top-two logit gap. The last uses no
labels.

\textbf{G, optimisation geometry,} on a fixed \emph{training} probe of 512
pairs, because gradients are what the optimiser sees: gradient norm,
gradient-to-weight ratio, cosine between gradient and weight vector, a
virtual AdamW update ratio formed from the optimiser state at the
checkpoint, and the dispersion of layer-wise gradient norms. The virtual
update includes both the adaptive term and the decoupled weight-decay term;
with $\lambda=1$ and learning rate $10^{-3}$ the decay term is the same
order as the adaptive term and differs between conditions.

\textbf{F, Fourier,} from the discrete transform of the embedding and
unembedding across the residue axis, with the DC component removed. These
are an oracle for modular arithmetic and are excluded from the general
representation $\Phi_{\mathrm{general}} = A \cup R \cup G$, which is
fifteen-dimensional.

All extraction runs on CPU in float64 with deterministic algorithms, and the
extractor verifies bitwise agreement across a repeated pass before
processing the cohort. The \textbf{conventional observables} $B_0$ are
weight norm, $\log_{10}$ training loss and $\log_{10}$ test loss. Test loss
is included because the endpoint is a test-loss difference: without it, a
descriptor that merely tracked the current test level would appear
informative about state.

\subsection{Analyses}

\textbf{Reproducibility of response.} The response is decomposed with cell,
condition, run, checkpoint and replicate components, and the estimand is
$\rho_{\mathrm{state}} = \sigma^2_{\mathrm{checkpoint}} /
(\sigma^2_{\mathrm{checkpoint}} + \sigma^2_{\mathrm{continuation}})$, with
seed-block and run effects removed so that a stable seed effect is not
counted as state signal. Estimation details, the exact projection used in
place of marginal double-demeaning, and the bootstrap and REML procedures
are given in Appendix~\ref{app:variance}.

\textbf{Residual information.} Each descriptor is regressed on a quadratic
basis in the three conventional observables. $R^2$ is out of sample by
leave-one-seed-block-out with the standardiser fitted inside each training
split; an in-sample $R^2$ would understate what $B_0$ explains and so
overstate the residual. A cubic basis is run as a sensitivity check.
Effective rank of the residual is the exponentiated entropy of the
normalised eigenspectrum.

\textbf{Within-seed alignment.} Descriptor residuals and the response
residual are each centred within seed block before testing, so shared seed
identity cannot register as alignment. Prediction is by
leave-one-seed-block-out ridge with an unpenalised intercept, swept over
five penalties. The null permutes the response in whole seed bundles within
availability classes --- sets of seeds occupying the same (cell, condition)
slots --- with values matched slot by slot so that nothing moves between
cells or conditions.

\textbf{Transport criterion.} For every cross-condition, cross-seed pair
within a cell, a baseline scorer uses the conventional observables and their
derivatives, and an augmented scorer retains every baseline predictor
unchanged and adds three descriptor-family distances. The ridge penalty is
selected using the baseline alone, by leave-two-seeds-out cross-validation
with both endpoints of an edge held out, and reused for the augmented model
without retuning. Each checkpoint selects its nearest partner under each
scorer; the statistic is the relative reduction in the mean absolute
response gap of the selected pairs, weighted equally across cells.

\textbf{Conditional placebo.} Each descriptor is decomposed as
$f(B_0) + \text{residual}$ by a fixed quadratic regression. The fitted part
is retained exactly; the residual is permuted as one fifteen-dimensional
block between seed bundles within availability classes, with a derangement
enforced rather than hoped for. The full pipeline, including baseline-only
penalty selection, is rerun for each permutation. The placebo therefore
preserves each descriptor's fitted dependence on $B_0$ and the residual
covariance, and removes only the alignment between internal geometry and
response. Its strength is reported as the correlation between placebo and
real family distances at the edge level.

\textbf{Positive control.} A known signal is injected: the response residual
after the $B_0$ fit, standardised, added to each descriptor's residual in
units of that descriptor's own residual standard deviation. The whole
pipeline is rerun at each level. Because injection inflates the residual
standard deviation, the detection threshold is converted through the same
measurement on the real residuals so injected and observed quantities share
a scale.

\textbf{Panel.} Three descriptor arms with exactly fifteen channels each:
the real descriptors; a synthetic set retaining the fitted quadratic
dependence on $B_0$ and replacing the residual with a Gaussian draw matched
to its covariance; and pure noise with no fitted dependence. Channel count
is held fixed to prevent representational dimensionality from becoming an
additional source of model capacity and variance reduction.

\section{Results}\label{sec:results}

\paragraph{Evidence regimes.} Two regimes, kept distinct throughout. Five
paired seed blocks were trained first and used to develop the design. Ten
fresh paired seed blocks were then trained, and R1 uses only those ten. The
descriptor and transport analyses in R2--R5 were developed and evaluated on
a pooled fifteen-block dataset comprising the original five and the ten used
in R1. Their leave-one-seed-block-out procedures therefore give out-of-block
prediction within that pooled dataset; they are not independent replication
on a study-level holdout.

\begin{table}[t]
\centering\small
\caption{A falsification ladder for training-state identification. Each
stage is read with its evidence regime, and each control removes one
alternative explanation for the stage above it.}
\label{tab:ladder}
\begin{tabular}{@{}lllll@{}}
\toprule
Stage & Question & Evidence unit & Outcome & Interpretation\\
\midrule
R1 & Is response reproducible? & fresh confirmatory runs &
$\rho_{\mathrm{state}} = \rhoAbstract$ & the target exists\\
R2 & Structure beyond $B_0$? & pooled development & residual rank \effRank{}
& residual structure exists\\
R3 & Tracks response locally? & within-run, seed-held-out & OOS $R^2$ up to
\oosRsqHi{} & local response relevance\\
R4 & Cross-run matching? & cross-run pairs & fails conditional placebo &
global geometry not established\\
R5 & Could R4 detect signal? & calibrated injection & separation at
$\alpha = \alphaCross$ & not explained by insensitivity\\
\bottomrule
\end{tabular}
\end{table}

\subsection{R1: Intervention response is a highly reproducible property of a
checkpoint}

\emph{Fresh confirmatory cohort: \nBlocks{} seed blocks, \nRuns{} runs.}

\textbf{Question.} If the same checkpoint is perturbed the same way twice,
does it respond the same way, and do different checkpoints differ by more
than that noise?

\textbf{Design.} \nCkpts{} checkpoints were selected by a mechanical rule,
frozen after development and before this cohort was trained, inside six
cells of a (weight norm, training loss) grid. Each was continued three times
with a weight-decay pulse and three times without, sharing random numbers
within a pair so the branches differ only in the intervention. The response
is the paired difference in $\log_{10}$ test loss at a fixed horizon.
Bitwise determinism was verified first: repeated control branches agree
exactly in model tensors, optimiser state and raw logits. The estimand is
$\rho_{\mathrm{state}} = \sigma^2_{\mathrm{checkpoint}} /
(\sigma^2_{\mathrm{checkpoint}} + \sigma^2_{\mathrm{continuation}})$, with
seed-block and run effects removed, so a stable seed effect is not counted
as state signal.

\textbf{Number.} At $h=500$, $\rho_{\mathrm{state}} = \rhoFull$, 95\% CI
$[\rhoCIpctLo, \rhoCIpctHi]$ by percentile and $[\rhoCIbcaLo, \rhoCIbcaHi]$
by BCa, from a cluster bootstrap over the ten seed blocks. A moment
estimator and a hierarchical REML fit agree to four decimals. Variance
components: $\sigma^2_{\mathrm{checkpoint}} = \sigmaCkpt$,
$\sigma^2_{\mathrm{continuation}} = \sigmaCont$. The ratio falls to
\rhoHone{} at $h=1000$ and \rhoHtwo{} at $h=2000$. Over that range the
checkpoint component grows, but the continuation component grows far faster.

\textbf{Interpretation.} At short horizons, intervention response is a
highly reproducible checkpoint-level property relative to continuation
variability. Its reproducibility declines with horizon as continuation
variance approaches checkpoint variance: what is lost is identifiability
through accumulated continuation uncertainty, not the checkpoint signal
itself. This establishes that intervention response is measurable as a
checkpoint-level target at the prespecified short horizon, which is what
makes an intervention-based notion of training state a coherent object to
look for.

\begin{figure}[t]\centering
\includegraphics[width=\linewidth]{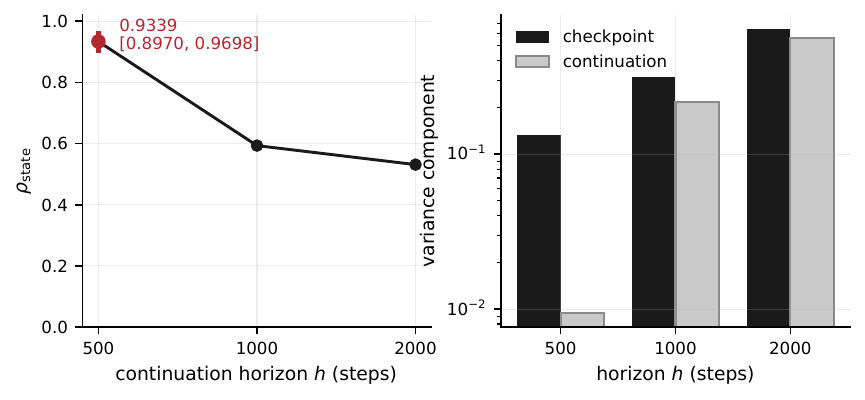}
\caption{Response reproducibility: variance
components and $\rho_{\mathrm{state}}$ across horizons ($\rhoFull$ at
$h{=}500$, BCa interval shown; \rhoHone{} and \rhoHtwo{} at longer
horizons).}\label{fig:fig1_reproducibility}
\end{figure}

\subsection{R2: Conventional observables explain most, but not all, of the
internal descriptors}

\emph{Development dataset: fifteen seed blocks, 216 checkpoints.}

\textbf{Question.} Fifteen general descriptors were computed from each
checkpoint's weight spectra, representation geometry and optimiser geometry.
Do they carry anything that weight norm, training loss and test loss do not
already determine?

\textbf{Design.} Each descriptor was regressed on a quadratic basis in the
three conventional observables, evaluated out of sample by
leave-one-seed-block-out with the standardiser fitted inside each training
split. A cubic basis was run as a sensitivity check against local
underfitting. Test loss is in the conditioning set because the response
endpoint is a test-loss difference; without it, a descriptor that merely
tracked the current test level would appear informative about state.

\textbf{Number.} Among the fifteen general descriptors, the median
out-of-sample $R^2$ given the conventional observables is \rTwoA{} for the
weight-spectrum family (A), \rTwoR{} for the representation family (R) and
\rTwoG{} for the optimiser-geometry family (G). Two retain substantially
more residual variation: gradient--weight cosine at \resCosine{} and the
dispersion of the top-two logit gap at \resLogitGap{}; the virtual AdamW
update ratio follows at \resAdamW{}. The residual of the fifteen has an
effective rank of \effRank{}; a cubic basis gives \effRankCubic{} and leaves
the ordering unchanged. The task-specific Fourier family (F), an oracle for
modular arithmetic and not part of the general representation, has a median
$R^2$ of \rTwoF{}.

\textbf{Interpretation.} Most general-descriptor variation is predictable
from the conventional observables, but a structured residual remains: the
residual covariance has effective rank \effRank{}, and several descriptors
retain substantial out-of-block variation, gradient--weight cosine and the
top-two logit gap most of all. The near-identical result at degree three
argues against local underfitting; it does not establish that the residual
is unpredictable by any non-linear learner, which was not tested. The
Fourier descriptors are, if anything, more predictable from norm and loss
than the general families are; their oracle status does not translate into
greater residual information beyond these conventional observables. Two of
the three least-determined general descriptors measure optimiser geometry,
which the conventional observables cannot see at all.

\begin{figure}[t]\centering
\includegraphics[width=\linewidth]{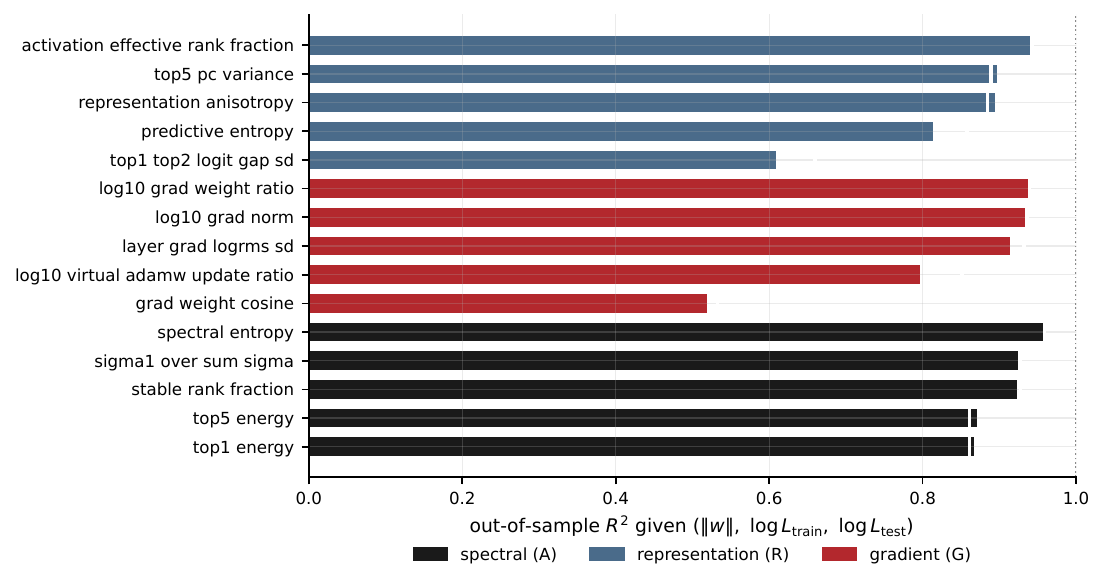}
\caption{Descriptor residual structure:
per-family out-of-sample $R^2$ given $B_0$ and the effective rank
(\effRank{}) of the residual.}\label{fig:fig2_residual_information}
\end{figure}

\subsection{R3: Within a run, descriptor deviations track response
deviations}

\emph{Development dataset: fifteen seed blocks, 148 checkpoints in five
cells.}

\textbf{Question.} Residual variance is not the same as relevant variance.
Does the part of the descriptors the conventional observables cannot predict
carry information about how the checkpoint will respond?

\textbf{Design.} Descriptor residuals and response were both centred within
seed block before testing, removing stable between-run offsets from both
sides. The estimand therefore asks whether checkpoint-level deviations from
a run's mean descriptor state track deviations from that run's mean
response. Prediction is by leave-one-seed-block-out ridge with an
unpenalised intercept. The null permutes the response in whole seed bundles
within availability classes, matched slot by slot, so a value never moves
between grid cells or between conditions.

\textbf{Number.} Out-of-sample correlation is \oosCorrLo{} to \oosCorrHi{}
across five ridge penalties, with out-of-sample $R^2$ from \oosRsqLo{} to
\oosRsqHi{}, clearing the seed-bundle permutation null at every penalty with
$p$ from \oosPLo{} to \oosPHi{}. Two of the fifteen seed blocks belong to
singleton availability classes and stay fixed under permutation, reducing
the randomisation available to separate the statistic from its null.

\textbf{Interpretation.} Within a training run, the residual descriptor
variation contains seed-held-out response-relevant information: the
descriptors know where a checkpoint sits relative to its own trajectory in a
way that bears on its future response. The multivariate alignment is
substantially stronger than any individual linear projection; a
per-descriptor analysis would have found little.

This is a seed-held-out association within the development dataset, not an
independent replication. The descriptor set, the conditioning variables and
the state-space construction were all developed using data that contribute
to this analysis. Leave-one-seed-block-out controls for overfitting to
individual runs; it cannot control for the choices that defined the
analysis.

\begin{figure}[t]\centering
\includegraphics[width=\linewidth]{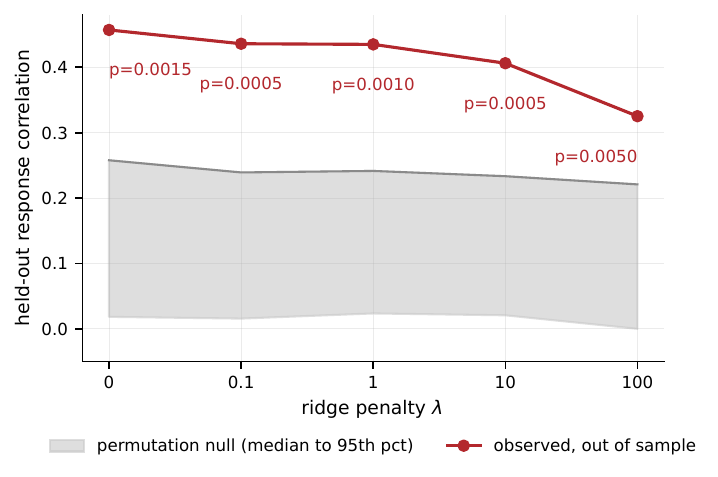}
\caption{Within-seed alignment: out-of-sample correlation
across ridge penalties against the seed-bundle permutation null.}\label{fig:fig3_within_seed_alignment}
\end{figure}

\medskip
R3 establishes a within-run result. Training-state equivalence, however, is
useful only when it can compare checkpoints whose training histories differ.
R4 therefore asks a deliberately stronger question: whether the same
representation supports response matching \emph{across} independently
trained seed blocks. R4 is not a second test of the R3 estimand. It tests
whether the within-run information identified in R3 organises the
representation in a way that transfers across runs.

\subsection{R4: Across runs, that information does not establish
equivalence}

\subsubsection{R4a: The apparent gain does not survive conditional
falsification}

\textbf{Question.} Given that the descriptors carry within-run
response-relevant information, do they identify checkpoints from
\emph{different} runs that will respond alike, under a transport criterion
fixed in advance?

\textbf{Design.} Every candidate pair joins two different seed blocks and
the two weight-decay conditions; same-run pairs are excluded by
construction. A baseline scorer uses the conventional observables and their
derivatives. The augmented scorer retains every baseline predictor unchanged
and adds three descriptor-family distances; the ridge penalty is selected
using the baseline alone and reused without retuning. Each checkpoint
selects its nearest partner under each scorer, and the scorers are compared
on the actual response gap of the pairs they chose. The statistic is the
relative reduction in that gap.

Retaining the baseline predictors does not by itself make the comparison
fair --- correlated additional channels can help refit the baseline surface
--- so the confound is addressed by the falsification rather than by the
model structure. Each descriptor is split into the part a quadratic function
of the conventional observables predicts and a residual; the fitted part is
kept exactly and the residual is permuted between seed bundles as one
fifteen-dimensional block.

\textbf{Number.} The observed gain is $G = \Greal$. Across 100 conditional
permutations the placebo median is \shamMedian{} and the 95th percentile
\shamQfive{}; \shamFrac{} of placebos reach or exceed the observed value.
The gain beyond the placebo median is \Gexcess{}.

The placebo remains structurally similar to the real descriptors rather than
degenerating into unrelated noise: its family distances correlate with the
real distances at Spearman \spearA{}, \spearR{} and \spearG{}. By
construction, the residual seed bundles are reassigned while the fitted
dependence on the conventional observables is retained.

\textbf{Interpretation.} An apparent \GrealPct{} improvement in matching
checkpoints across runs does not separate from a control that preserves the
descriptors' fitted dependence on the conventional observables. The
within-run information established in R3 does not, by itself, deliver a
cross-run response-equivalence geometry under this transport criterion.

The placebo distribution is broad, which raises the possibility that this
failure reflects limited sensitivity rather than absent transportable
structure. R5 tests that explanation directly.

\subsubsection{R4b: The placebo background tracks the link to the
conventional observables}

\textbf{Question.} Why is the placebo's own gain around
\panelMedARGShort{} rather than zero?

\textbf{Design.} Three descriptor arms, each with exactly fifteen channels:
the real descriptors; a synthetic set retaining the fitted quadratic
dependence on the conventional observables and replacing the residual with a
Gaussian draw matched to its covariance; and pure noise with no fitted
dependence. Each arm ran through the same pipeline against its own placebo.
The Fourier family was excluded: it has five channels and cannot be compared
with a fifteen-channel arm without fabricating columns.

\textbf{Number.} The placebo median is \panelMedARG{} for the real
descriptors and \panelMedCorr{} for the correlation-matched synthetic ---
effectively identical --- while the pure-noise arm's placebo median falls to
\panelMedNoise{}. The synthetic arm's own observed gain is \panelRealCorr{},
well below the real \Greal{} and below its own placebo median.

\textbf{Interpretation.} Preserving fitted dependence on the conventional
observables reproduces the ${\approx}\panelMedARGShort$ placebo background;
removing that dependence reduces the background to
${\approx}\panelMedNoiseShort$. But a covariance-matched synthetic
descriptor does not reproduce the full observed gain, so the real
descriptors carry structure the synthetic does not. That extra structure is
what fails the falsification in R4a. Correlation with the conventional
observables therefore accounts for the background, not for the whole
observed gain --- and the remainder still does not survive.

\begin{figure}[t]\centering
\includegraphics[width=\linewidth]{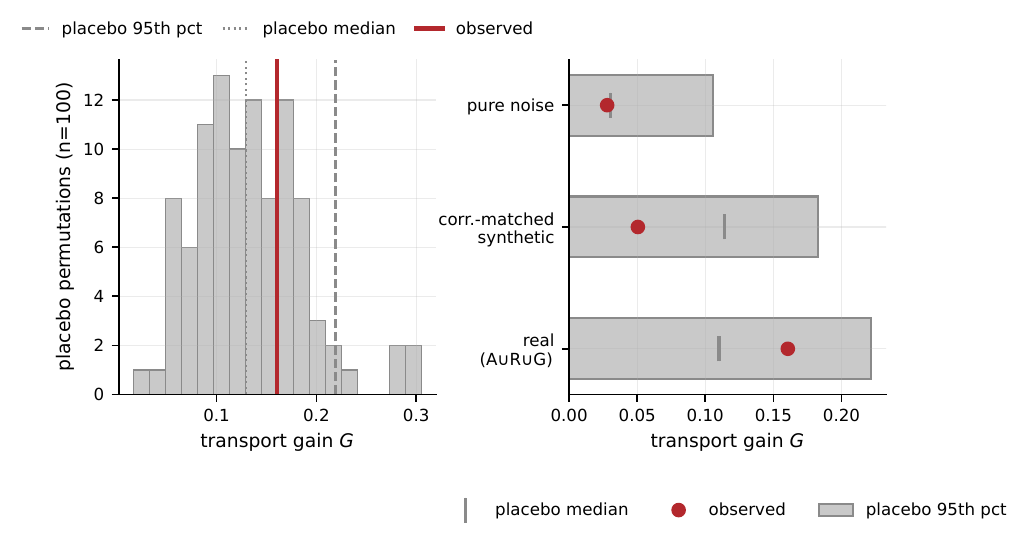}
\caption{Conditional falsification.
(a)~Primary sham: observed $G=\Greal$ against the conditional placebo
distribution (median \shamMedian{}, 95th percentile \shamQfive{}).
(b)~Placebo panel: real, correlation-matched synthetic, and pure-noise
arms; placebo backgrounds \panelMedARG{}, \panelMedCorr{},
\panelMedNoise{}.}\label{fig:fig4_placebo_falsification}
\end{figure}

\subsection{R5: What the falsification could have detected}

\emph{Development dataset: fifteen seed blocks, 148 checkpoints in five
cells.}

\textbf{Question.} A negative result is informative only if the test could
have detected a positive one. How much signal does this falsification need?

\textbf{Design.} A known signal was injected: the part of the response the
conventional observables cannot predict, standardised, added to each
descriptor's residual in units of that descriptor's own residual standard
deviation. At $\alpha=0$ this is the real descriptor. The entire pipeline
--- fit, placebo, comparison --- was rerun at each level. Because injection
inflates the residual standard deviation, the detection threshold was
converted through the same measurement on the real residuals, so injected
and observed quantities share one scale.

\textbf{Number.} Detection crosses at $\alpha=\alphaCross$: add-one
exceedance falls from \excHalf{} at $\alpha=0.5$ to \excOne{} at
$\alpha=\alphaCross$ and stays there. In the same units, the real
descriptors' aligned component reads \alphaEffMed{} against a threshold of
\alphaThresh{}.

\textbf{Interpretation.} Under the injected linearly aligned signal model,
separation from placebo becomes reliable once the injected residual-scale
signal reaches $\alpha=\alphaCross$ --- \alphaThresh{} against the
descriptors' own \alphaEffMed{} on the calibrated $\alpha_{\mathrm{eff}}$
scale, about \ratioFold-fold --- and weaker signals need not be detected.
The R4 outcome is therefore not readily explained by insensitivity to a
linearly aligned signal at or above that scale, and the paper claims nothing
about signals below it.

The calibration constrains sensitivity to the signal class injected.
Structure expressed non-linearly, or distributed across directions the
projection does not capture, lies outside it.

\begin{figure}[t]\centering
\includegraphics[width=\linewidth]{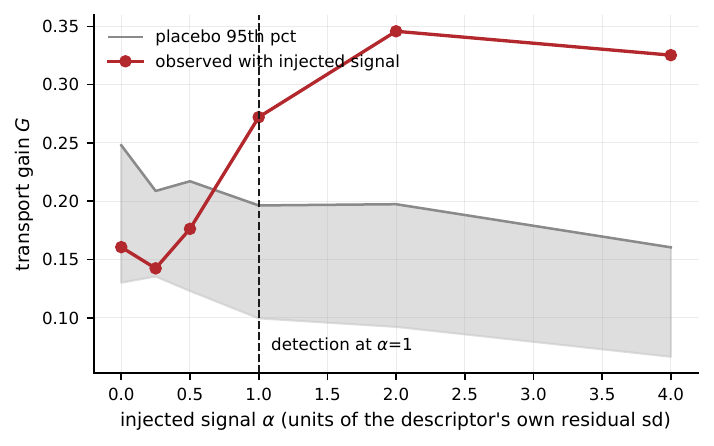}
\caption{Positive-control calibration: observed
$G(\alpha)$ at $\alpha \in \{0, 0.25, 0.5, 1, 2, 4\}$ =
\{\GcurveZero{}, \GcurveQuarter{}, \GcurveHalf{}, \GcurveOne{},
\GcurveTwo{}, \GcurveFour{}\} against per-level placebo 95th percentiles;
the curve is non-monotonic and is drawn as measured.}\label{fig:fig5_calibration}
\end{figure}

\section{Discussion}\label{sec:discussion}

The central finding is not that the descriptors fail to encode training
dynamics. They do so locally, within a run: after removing stable run
effects from both sides, their residual variation predicts
residual intervention response across held-out seed blocks, across all five
reported ridge penalties, against a permutation null constrained to respect
the design. What fails is the stronger requirement that these locally informative
coordinates align across independently trained trajectories well enough to
support cross-run state-equivalence judgments.

\subsection{Predictability and equivalence are different requirements}

The two halves are not in conflict, and seeing why is the point.

Prediction is a property of the representation \emph{together with a
decoder}: it asks whether some function of the descriptors recovers the
attribute. R3 establishes exactly that and no more --- a linear combination
of fifteen descriptors, fitted without each held-out seed block in turn,
predicts response deviation in that block.

Equivalence judgments depend on the \emph{pairwise geometry} the
representation induces. They ask whether that geometry orders pairs by
similarity of future response. A representation can contain a direction
along which response varies and still place two checkpoints with very
different futures close together, because the distance deciding nearness is
dominated by directions carrying no response information.

Formally, the two ask different things. Prediction asks whether there exists
$f$ with $f(\Phi(x)) \approx y$. Equivalence asks whether
$d_\Phi(x_i, x_j)$ orders pairs consistently with $d_y(y_i, y_j)$. The
existence of $f$ does not imply the second property, and this paper is a
concrete case where it does not.

The failure is therefore not a failure to encode response; it is a failure
of that encoded information to support the response-equivalence judgments
required by our transport criterion.

The experiments separate two axes that are often conflated: whether
information predicts an attribute and whether it organises pairwise
similarity, and whether either property is local to a training trajectory or
aligned across independently trained trajectories. We measure one cell of
that pairing on each axis and leave the others open.

\subsection{Local charts and a global coordinate}

The within-run and cross-run results sit in a hierarchy that makes the
pattern legible. Future response is a checkpoint-specific and reproducible
property: this is established on fresh runs. The descriptors locate a
checkpoint relative to its own trajectory in a way that bears on response:
this is positive development evidence. Whether the same descriptors let us
say that a checkpoint from one trajectory and a checkpoint from another
occupy the same state: this is not established.

Each trajectory may admit a serviceable local coordinate without those
coordinates being aligned into a common system. That is a weaker and more
specific statement than ``the descriptors carry no state information'', and
it is what the evidence supports.

\subsection{Where the apparent advantage comes from}

The placebo distribution is not centred at zero, and understanding why is
what makes the negative result interpretable rather than merely
disappointing.

Permuting the descriptors' residual while retaining their fitted dependence
on the conventional observables leaves a placebo background near
\panelMedARGShort{} in the panel analysis. A synthetic descriptor with no
measured content but the same fitted dependence produces almost exactly the
same placebo background, \panelMedCorrShort{}; a descriptor with no fitted
dependence at all has a placebo background of only \panelMedNoiseShort{}.
The placebo background therefore tracks the retained fitted dependence on
the conventional observables.

The synthetic controls are consistent with a mechanism in which correlated
auxiliary channels improve the approximation of structure already available
through the baseline, even in the absence of additional response-specific
content. This identifies a confound that can arise when richer
representations are evaluated against leaner conventional baselines; it does
not imply that every such comparison is affected to the same degree.

The real descriptors are not merely the conventional observables in
disguise: their observed gain of \GrealShort{} exceeds both the background
and the \panelRealCorrShort{} achieved by the correlation-matched synthetic,
so they carry structure the synthetic does not. What the evidence says is
narrower and harder to escape --- that additional structure was not enough
to separate them from their own conditional placebo.

\subsection{What follows}

If passive descriptors of a checkpoint can carry within-run
response-relevant information without furnishing a cross-run equivalence
geometry, one alternative is to characterise state through responses
themselves. A vector of reproducible responses to several controlled
training interventions would be an active phenotype rather than a static
description, and its value would be tested by predicting the response to an
intervention not used to construct it. Our first result is a precondition
for that programme; we have not run it, and nothing here shows it would
succeed.

The results suggest that training state may be easier to identify by how a
network responds than by how it looks.

\section{Limitations}\label{sec:limitations}

\textbf{One task, one architecture.} Modular addition with $p=97$, a
one-layer transformer, AdamW. Whether any of this transfers is untested, and
the reproducibility result in particular may depend on the unusually
structured dynamics of grokking systems.

\textbf{Equivalence is relative to an intervention and a horizon.} The
definition is stated with respect to a weight-decay pulse, and every
descriptor analysis uses the prespecified $h=500$ response, where the
confirmatory experiment established the highest reliability. Conclusions
about descriptor relevance and transport apply at that horizon and are not
extrapolated to the longer ones, where $\rho_{\mathrm{state}}$ falls to
\rhoHoneShort{} and \rhoHtwoShort{}.

\textbf{The within-run positive result has not been independently
replicated.} It was obtained under seed-held-out prediction within the
pooled development dataset. Because the descriptor set, the conditioning
variables and the state-space construction were informed by data in that
pool, leave-one-seed-block-out controls for overfitting to individual runs
but not for the choices that defined the analysis. We treat it as motivating
rather than establishing a general response-relevant representation.

\textbf{The cross-run failure was observed on that same dataset, under one
specification.} As a negative result it is less exposed to the particular
selection-induced optimism affecting the within-run finding, but it is not
thereby conservative and has not been tested beyond this task, architecture,
and descriptor set.

\textbf{The placebo's validity is empirical, not inherited.} It follows the
conditional-randomisation principle but is not one of the exact procedures
for which guarantees exist: a fitted low-order conditional, seed-bundle and
availability constraints, and a statistic that is the output of a downstream
fitting-and-selection pipeline. We calibrate its detection threshold and
report that calibration. The calibration constrains sensitivity to the
linearly aligned signal class injected; non-linear structure, or signal
distributed across directions the projection does not capture, remains
outside it.

\makeatletter
\subsubsection*{Reproducibility statement}
Every reported number resolves to a canonical script, its inputs and a
frozen artifact, each identified by SHA-256 in a provenance manifest; the
figures are generated from those artifacts by a single script that writes
the exact plotted values to companion CSV files, and headline numbers enter
the manuscript only through a shared macro file, never by hand.
\if@accepted
Code, the frozen artifacts, the manifest and the figure pipeline are
available at \repourl.
\else
An anonymized reproducibility package containing the scripts, frozen
artifacts, manifest and figure pipeline accompanies this submission as
supplementary material.
\fi

\if@accepted
\ifx\ackstext\@empty\else
\subsubsection*{Acknowledgments}
\ackstext
\fi
\else
\subsubsection*{Acknowledgments}
Withheld for double-blind review.
\fi
\makeatother

\bibliography{references}
\bibliographystyle{tmlr}

\appendix

\section{Checkpoint selection}\label{app:selection}

Checkpoints are chosen by a rule fixed before the confirmatory cohort was
trained, so that no checkpoint enters the analysis because of how it
responded.

Each run is first scanned at a resolution of 100 steps, recording step,
training loss, training and test accuracy, weight norm, and a hash of the
model state. No checkpoint is written during the scan.

A $3\times 3$ grid is then defined on (weight norm, $\log_{10}$ training
loss) by a mechanical rule: restrict to scan points strictly before each
run's \emph{first passage} of test accuracy 0.95; discard points inside the
initialisation neighbourhood, defined as those whose weight norm is within
5\% of the run's initial norm; and take tertiles of the pooled weight norm
and of the pooled $\log_{10}$ training loss over what remains. Bins are
half-open, $[\mathrm{left}, \mathrm{right})$. The rule consults trajectory
geometry only; it never consults response.

First passage rather than an instantaneous threshold matters: runs re-cross
0.95 downwards long after grokking, and an instantaneous test would readmit
late checkpoints as though they were pre-transition. On the confirmatory
cohort this distinction reclassifies 47 scan rows.

Within each grid cell, each run contributes the single scan point closest to
the cell centre in cell-normalised (weight norm, $\log_{10}$ loss)
coordinates. Distance is computed in the same geometry the grid was defined
in --- log for the loss axis --- because a linear midpoint of a bin spanning
two orders of magnitude sits near its upper edge and would select
systematically higher-loss checkpoints.

Two support conditions, fixed before the confirmatory cohort, decide which
cells enter the analysis: every condition must contribute at least three
seed blocks to a cell, and at least three cells must satisfy this. On the
confirmatory cohort six of nine cells qualify, yielding 145 checkpoints, of
which \nCkpts{} lie in the qualifying cells.

Selected checkpoints are then \emph{rematerialised}: each run is retrained
deterministically from its seed and the model state is written only at the
selected steps, with the model hash, training loss and weight norm checked
against the scan at each. A mismatch aborts the run rather than producing a
checkpoint.

\section{Determinism enforcement}\label{app:determinism}

cuDNN and cuBLAS are set to deterministic algorithms,
\texttt{CUBLAS\_WORKSPACE\_CONFIG} is set before torch is imported, and
\texttt{torch.use\_deterministic\_algorithms(True)} is enabled without
\texttt{warn\_only}. Before measurement, a control branch is run twice from
the same seed on three checkpoints spanning early, middle and late training,
and the two runs are required to agree bitwise in model tensors, optimiser
state and raw logits. Without this, kernel nondeterminism would enter the
paired difference, which is the quantity being measured.

Restoring a state dictionary can preserve storage pointers, so an optimiser
step in one branch can mutate the checkpoint the other branch is about to
read. Every branch asserts that no tensor in its model or optimiser state
shares storage with the checkpoint it was restored from.

\section{Variance estimation for R1}\label{app:variance}

The variance decomposition is
\begin{equation*}
D_{ikr} \;=\; \mu + \alpha_{\mathrm{cell}} + \gamma_{\mathrm{condition}}
+ u_{\mathrm{run}} + v_{\mathrm{checkpoint}} + \varepsilon_{\mathrm{replicate}},
\end{equation*}
and the estimand is $\rho_{\mathrm{state}} = \sigma^2_v / (\sigma^2_v +
\sigma^2_\varepsilon)$. The run and seed-block components are removed and
are not counted as state signal: two checkpoints from one run resembling
each other because they share a history is not what the claim is about.

$\sigma^2_\varepsilon$ is the pooled within-checkpoint variance of $D$
across replicates. $\sigma^2_v$ is estimated from the checkpoint means after
removing cell and run effects by an exact least-squares projection --- not
by marginal double-demeaning, which coincides with it only for a balanced
design, and this design is not balanced --- and corrected for the sampling
error of each checkpoint mean.

Intervals come from a cluster bootstrap over the ten seed blocks, reported
as percentile and BCa. A hierarchical REML fit with seed block as grouping
factor and checkpoint nested within run is reported as a sensitivity
analysis.

Because each run contributes at most one checkpoint per cell, the checkpoint
effect within a cell is not separable from a run $\times$ cell interaction.
This is the comparison of interest --- different runs at the same state
coordinate --- but the hierarchy is two-level, not three, and is described
as such.

\section{Software, determinism and provenance}\label{app:provenance}

PyTorch 2.11 with CUDA 12.8 on a single workstation GPU; analyses in NumPy,
pandas and statsmodels. Each experimental unit is written as an atomic file
and every stage resumes from what is already on disk, so an interrupted run
loses at most one unit. Intermediate traces --- per-continuation loss,
accuracy and weight-norm trajectories, raw logits at horizons, and per-unit
permutation records --- are retained rather than deleted.

A manifest maps each reported result to its canonical script, its inputs and
its artifact, with a SHA-256 for each, and lists the scripts and artifacts
that were superseded so that an obsolete file cannot be mistaken for a live
one. Headline numbers in the manifest are read from the artifacts rather
than transcribed. After the manifest was frozen, no new inferential analysis
of the modular-addition data was introduced. The frozen residual-information
artifact additionally records that the descriptor analysis is
response-blind: descriptors were extracted and modelled without access to
the response. The SHA-256 of the frozen manifest is
\texttt{\manifestHash}.

\section{Design development}\label{app:development}

The analysis was developed on the pilot cohort before evaluation on fresh
runs. Several design choices were revised during development, including the
replacement of an ineffective hard-caliper formulation by continuous
conditioning, and the addition of current test loss to the conventional
conditioning set. Implementation errors found during auditing were corrected
before the reported analyses were frozen.

We therefore describe the final pipeline as development-informed rather than
fully preregistered. The complete amendment history, the superseded
analyses, and the provenance hashes are given in the released manifest.

\end{document}